\documentclass[final]{cvpr}

\usepackage{times}
\usepackage{epsfig}
\usepackage{graphicx}
\usepackage{amsmath}
\usepackage{amssymb}
\usepackage{cite}
\usepackage[pagebackref=true,breaklinks=true,colorlinks,bookmarks=false]{hyperref}
\usepackage{subfigure}
\usepackage[export]{adjustbox}
\usepackage{makecell}
\usepackage{array}

\def\cvprPaperID{13}
\def\confYear{CVPR 2021}

\newif\ifsqueeze
\squeezefalse  
\ifsqueeze

\else

\fi

\begin{document}

\title{Image-to-Image Translation of Synthetic Samples for Rare Classes} 

\author{Edoardo Lanzini\\
Delft University of Technology\\
Mekelweg 5, 2628 CD Delft, Netherlands\\
{\tt\small e.m.lanzini@student.tudelft.nl}
\and
Sara Beery\\
California Institute of Technology\\
1200 E California Blvd, Pasadena, CA 91125\\
{\tt\small sbeery@caltech.edu}
}

\maketitle

\begin{abstract}
The natural world is long-tailed: rare classes are observed orders of magnitudes less frequently than common ones, leading to highly-imbalanced data where rare classes can have only handfuls of examples.
Learning from few examples is a known challenge for deep learning based classification algorithms, and is the focus of the field of low-shot learning. One potential approach to increase the training data for these rare classes is to  augment the limited real data with synthetic samples. This has been shown to help, but the domain shift between real and synthetic hinders the approaches' efficacy when tested on real data.
 
 We explore the use of image-to-image translation methods to close the domain gap between synthetic and real imagery for animal species classification in data collected from camera traps: motion-activated static cameras used to monitor wildlife. We use low-level  feature alignment between source and target domains to make synthetic data for a rare species generated using a graphics engine more ``realistic". Compared against a system augmented with unaligned synthetic data, our experiments show a considerable decrease in classification error rates on a rare species.
 
\end{abstract}

\section{Introduction}
Accurately and scalably monitoring biodiversity is vital to our understanding of the changing world around us. Policymakers need near-real-time monitoring data to analyze the efficacy of conservation actions in the face of human encroachment and climate change. Camera traps and other static passive monitoring sensors provide vital monitoring data to ecologists, but as the size of these networks of sensors increase, the magnitude of data outpaces human processing capacity. Ecologists are increasingly turning to computer vision and machine learning approaches to help automate the detection and categorization of animal species, necessary in order to scale this critical assessment.

\begin{figure}[h]
\begin{center}
\begin{tabular}{m{0.5cm}>{\centering}m{3.25cm}>{\centering}m{3.25cm}c}
  &  \textit{day} &  \textit{night} &\\
\textit{real} & \includegraphics[width=.8\linewidth]{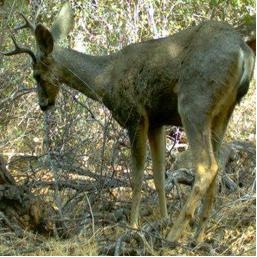} & \includegraphics[width=.8\linewidth]{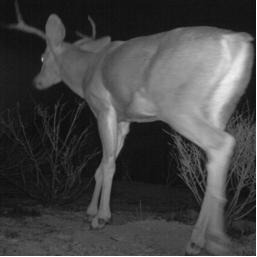} &\\
\textit{syn} & \includegraphics[width=.8\linewidth]{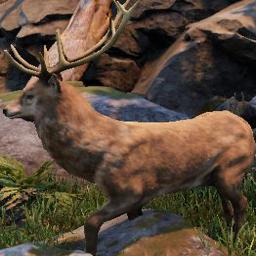} & \includegraphics[width=.8\linewidth]{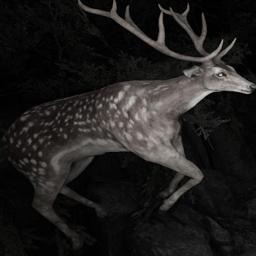} &\\
\end{tabular}
\end{center}
\caption{\textbf{Examples of real and synthetic images of deer.} The visual difference between the two domains is noticeable for both \textit{day} and \textit{night} examples.}
\label{fig:boxes}
\end{figure}

Camera trap data introduces challenges beyond those addressed in traditional computer vision benchmark datasets like ImageNet \cite{deng2009imagenet}. These include long-tailed distributions \cite{van2017devil} and a multitude of different sub-domains (locations) within the same dataset \cite{Beery_2018_ECCV}. In particular, the classification of rare species of animals is notoriously troublesome due to the combined effect of scarcity in number of examples and the low sample efficiency of data from a given camera deployment. 

To limit the bias toward well-represented classes, both algorithmic  \cite{elkan2001foundations, he2008adaptive, he2009learning} and data solutions \cite{bondi2018airsim} have been proposed. Beery \textit{et al.} explored the addition of synthetic samples for a single rare class and showed to improve classification accuracy \cite{Beery_2020_WACV}. However, despite the impressive capabilities of graphical engines, the synthetic samples are still perceived by the network as semantically distant compared to the real ones \cite{Beery_2020_WACV}.

Beery \textit{et al.} crafted a dataset starting from the Caltech Camera Traps (CCT) Dataset \cite{Beery_2018_ECCV}, artificially undersampling the deer class and training the classification model with synthetic renderings \cite{Beery_2020_WACV}. The same synthetic data is used as a starting point in this work to investigate the impact of synthetic-to-real image-to-image translation on the classification of the single rare class.

In this work we quantify the domain shift between synthetic and real camera trap data using color distribution, texture, and feature distance. We narrow the gap with unpaired image-to-image translation methods operating in a low-data regime with only a handful of real samples from the target domain. We show this results in higher efficacy when using synthetic data to augment limited real examples for a rare species, ultimately leading to an increase in classification performance for both seen and unseen locations.

\section{Related work}

\subsection{Domain Adaptation from Synthetic to Real}
Domain adaptation techniques often operate in the feature space, seeking to close the distribution gap between samples from different domains \cite{saenko2010}. Supervised and unsupervised techniques are used to align the features of the source (synthetic) and the target (real) \cite{cycada, visda, liu2017unsupervised, bousmalis2017}.
The gap is commonly bridged by either mapping the two domains to a domain-invariant representations \cite{dabackprop, ganin2016domain} or forcing the two learned distributions to be close \cite{gretton2012kernel, sun2016return, sun2016deep}. Various metrics have been proposed to measure this domain gap, including maximum mean discrepancy \cite{longwang}, correlation distance \cite{coral}, or adversarial discriminator accuracy \cite{dabackprop, datzeng}. 
Hoffman \textit{et al.} introduced Cycle-Consistent Adversarial Domain Adaptation (CyCADA), operating at both pixel and feature-level, showing significant improvements over previous methods \cite{cycada}. 

\subsection{Image-to-Image Translation}
As an alternative to feature-level domain adaptation, image-to-image (I2I) translation attempts to directly increase the ``realism" of synthetic data at the pixel-level. Paired I2I \cite{isolapix2pix} maps an image from source to target domain using an adversarial loss \cite{gans}, combined with a reconstruction loss between the result and target. In the unpaired setting, the samples from the two domains are not paired, and correspondence is enforced using cycle-consistency \cite{CycleGAN2017, kim2017learning, yi2017dualgan}, learning the mapping in both directions and computing a loss on the reconstruction of the original input.

Early adaptations of Generative Adversarial Networks (GANs) \cite{gans} showed promising results in simple settings, with small images and minimal semantic difference between domains \cite{shrivastava2017learning, coupledgans, bousmalis2017}. 

CycleGAN \cite{CycleGAN2017} uses a cycle consistency loss in its adversarial approach, training two different generators to translate in opposite directions, introducing a reconstructions loss. The architecture introduced by Isola \textit{et al.} is often extended with context-specific loss terms that allows to enforce further constraints on the translation learned \cite{ptgan, gazegan}.

UNIT \cite{liu2017unsupervised}, used in this work, is an I2I framework based on Coupled GANs \cite{coupledgans}. Compared to CycleGAN, the network does not learn a direct mapping between the two domains but instead operates under the assumption of a common latent space, in which both domains can be mapped. This assumption also implies a cycle-consistency constraint between the two domains \cite{liu2017unsupervised}. The adversarial setting of both UNIT and CycleGAN makes training complex. 

Recent work by Park \textit{et al.}, tackles the unpaired I2I problem using contrastive learning, operating at the level of patches, enforcing the constraint that corresponding patches in the two domains should have high mutual information \cite{CUT}. This intuition is formulated using a multilayer, patchwise contrastive loss that allows to learn a one-sided translation.

The application of I2I to translate from synthetic to real has improved performance in other real-world applications \cite{peppers, shrivastava2017learning, atapattu2019improving, liu2020stereogan, doersch2019sim2real}.

\subsection{CV for Camera Trap Data}
Camera traps are increasingly used by biologists to unobtrusively monitor wildlife. The use of deep learning to increase data processing speeds has been widely investigated in recent years \cite{cameratrapml2018, activelearning2021, schneider2019past, beery2018iwildcam, beery2019iwildcam, beery2020iwildcam, beery2019efficient, tabak2020improving, beery2020context, schneider2020}. The static nature of camera traps, combined with the long-tailed distribution of species in the real world, leads to poor generalization performance in novel deployments and for rare species \cite{schneider2020, Beery_2018_ECCV, WILDS2020}. Recent works tackle these challenges directly, focusing on categorization of rare species or generalization to novel camera deployments. Beyond data augmentation approaches like the one explored in this work, architectures \cite{longtail2021} and loss functions \cite{cui2019class, lin2017focal} designed for  long-tailed distributions have also shown promise.

\begin{figure}[t]
\begin{center}
\includegraphics[width=1\linewidth]{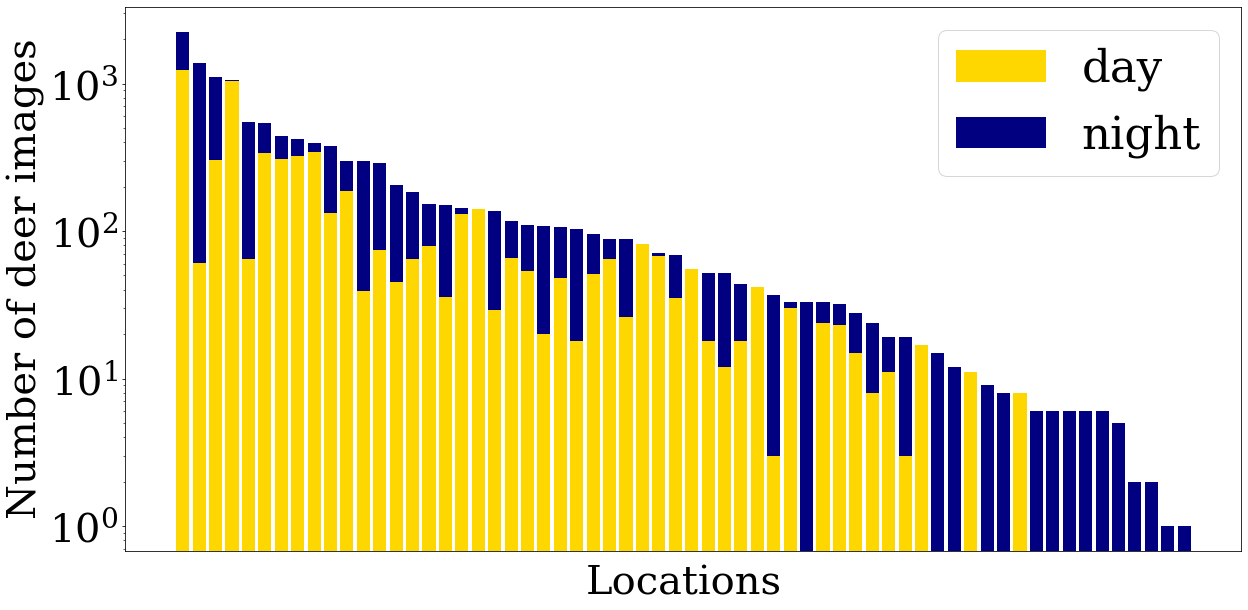}
\end{center} \caption{\textbf{Distribution of deer across CCT locations.} The number of deer seen at different camera locations is long-tailed and often combined with an uneven split between \textit{day} and \textit{night} within the same location. Note that the y axis is in log scale.}
\label{fig:longtail}
\end{figure}

\begin{figure*}
\begin{center}
\begin{tabular}{ c c c c }
\textbf{day} &
\includegraphics[width=.25\linewidth,valign=m]{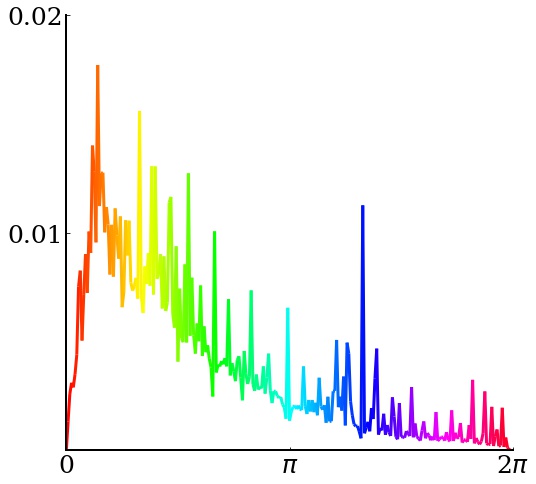} &
\includegraphics[width=.25\linewidth,valign=m]{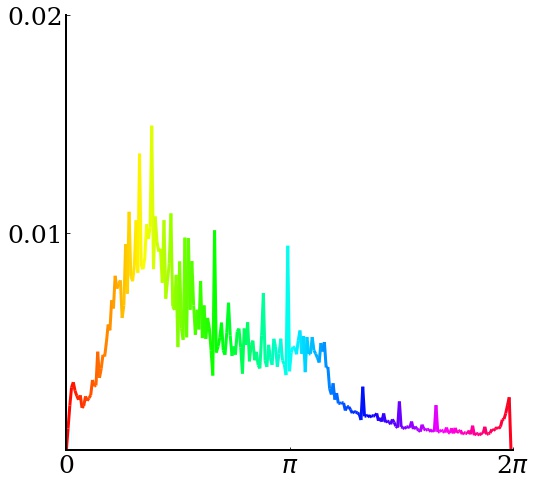} &
\includegraphics[width=.25\linewidth,valign=m]{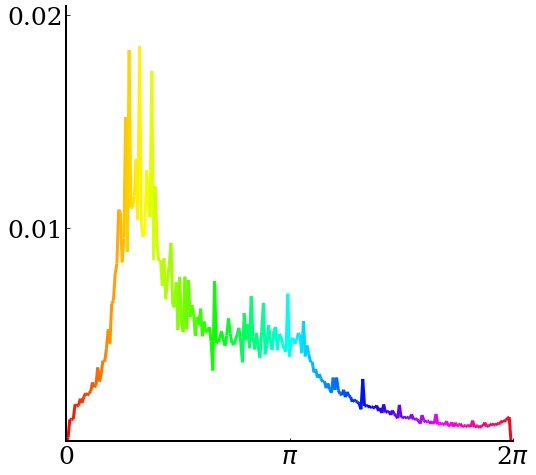} \\[+2ex]

\textbf{night} &
\includegraphics[width=.25\linewidth,valign=m]{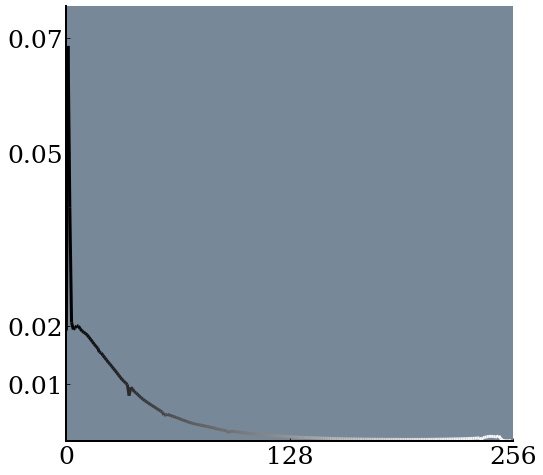} &
\includegraphics[width=.25\linewidth,valign=m]{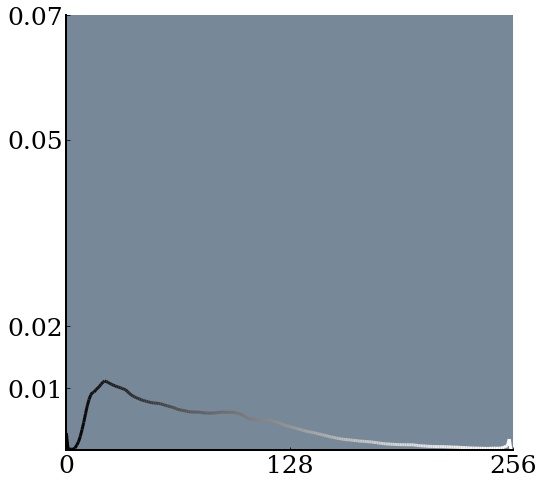} &
\includegraphics[width=.25\linewidth,valign=m]{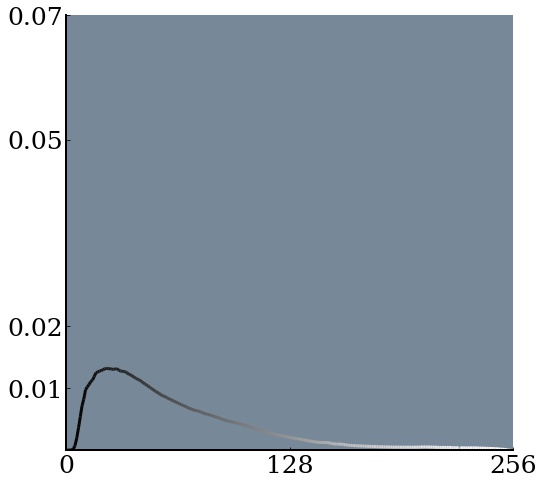} \\[+2ex]

& \textit{syn} & \textit{real} & \textit{syn2real}\\[+2ex]
\end{tabular}
\end{center}
\caption{\textbf{Comparison between the \textit{syn}, \textit{real} and \textit{syn2real} color distribution for day (top) and night (bottom).} The color distribution for the day is computed aggregating the discretized hue channel. For the night, samples are first converted to grayscale and then pixel values are aggregated. The resulting distribution of both is normalized. The \textit{syn2real} distributions move closer to the \textit{real} ones compared to the \textit{syn}. The high intensity of \textit{syn} night is due to the high saturation of the renderings.}
    \label{fig:hsv_combined}
\end{figure*}

\section{Data}

\subsection{CCT}
The Caltech Camera Traps (CCT) dataset contains 243,187 images from 140 camera trap locations covering 30 classes of animals, curated from data provided by the United States Geological Survey and the National Park Service \cite{Beery_2018_ECCV}. CCT is used as a testbed for long-tailed distributions under real-world conditions, where the number of samples for each species is unbalanced. The distribution of samples per-sensor is also long-tailed, with an additional uneven split between day and night occurrences (see Fig. \ref{fig:longtail}).

\subsection{CCT-20}
We use the same data split as \cite{Beery_2020_WACV}, starting with the CCT-20 subset introduced in \cite{Beery_2018_ECCV} and isolating deer as the single rare class of interest. The \textit{real} training set is composed of 13,553 images from 9 camera locations, containing only 44 deer examples, and is used as the source of real samples for our I2I translation task. Our additional \textit{synthetic} training data is also the same as \cite{Beery_2020_WACV}, which is generated with Unity's 3D game development engine. To constrain the task of translation, we make use of the bounding box annotations for both real and synthetic data to build two sets of images that share similar framing (see Figure \ref{fig:boxes}).

\begin{figure*}
\begin{center}
\centering
\begin{tabular}{ c c c c c }
\textit{syn} &
\includegraphics[width=.172\linewidth,valign=m]{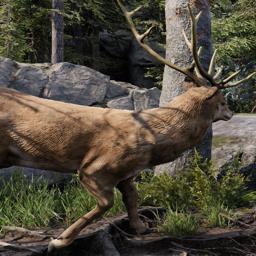} &
\includegraphics[width=.172\linewidth,valign=m]{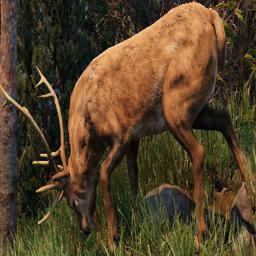} & 
\includegraphics[width=.172\linewidth,valign=m]{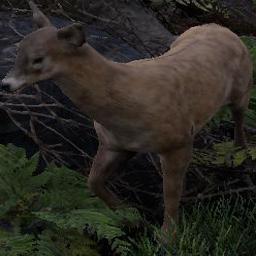} &
\includegraphics[width=.172\linewidth,valign=m]{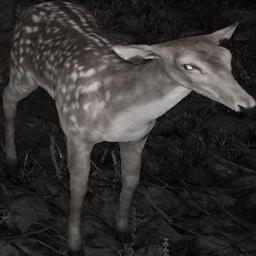} \\[+2ex]

\textit{syn2real} &
\includegraphics[width=.172\linewidth,valign=m]{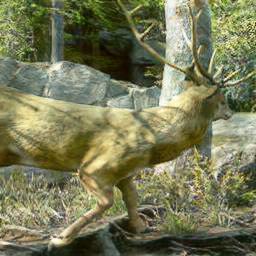} &
\includegraphics[width=.172\linewidth,valign=m]{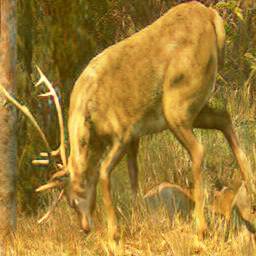} & 
\includegraphics[width=.172\linewidth,valign=m]{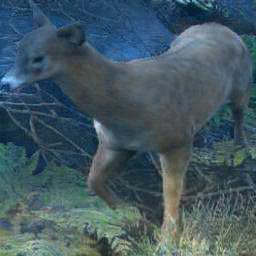} &
\includegraphics[width=.172\linewidth,valign=m]{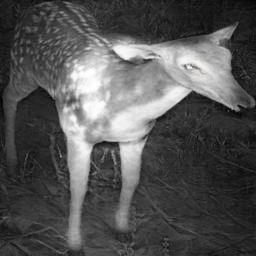} \\[+2ex]
\textit{real} & 
\includegraphics[width=.172\linewidth,valign=m]{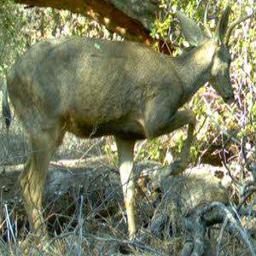} &
\includegraphics[width=.172\linewidth,valign=m]{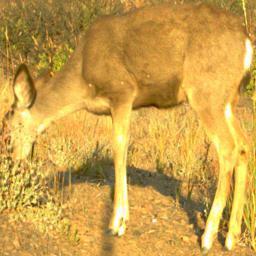} & 
\includegraphics[width=.172\linewidth,valign=m]{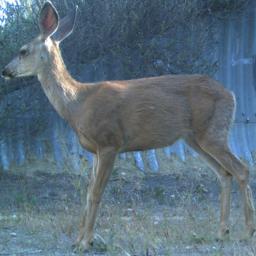} &
\includegraphics[width=.172\linewidth,valign=m]{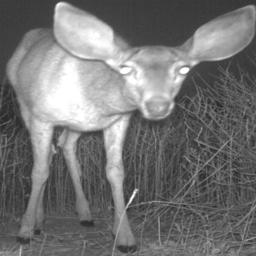} \\[+2ex]
\end{tabular}
\end{center}
\caption{\textbf{Examples generated by UNIT trained on CCT-deer.} These hand-picked examples show deer in similar poses starting with the \textit{syn} and comparing the two outputs of the models (\textit{syn2real}) with the \textit{real} sample.  The translation learns to match the color distribution of the real imagery, while the texture appears unchanged.}
\label{fig:deer_matrix}
\end{figure*}

\begin{figure*}
\begin{center}
\centering
\begin{tabular}{ c c c c c }
\textit{syn} &
\includegraphics[width=.172\linewidth,valign=m]{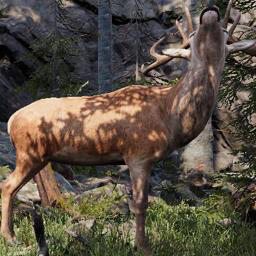} &
\includegraphics[width=.172\linewidth,valign=m]{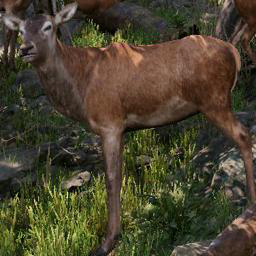} & 
\includegraphics[width=.172\linewidth,valign=m]{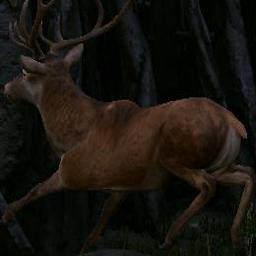} &
\includegraphics[width=.172\linewidth,valign=m]{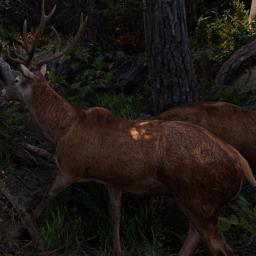} \\[+2ex]

\textit{syn2real} &
\includegraphics[width=.172\linewidth,valign=m]{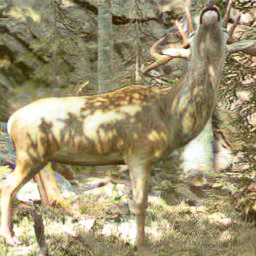} &
\includegraphics[width=.172\linewidth,valign=m]{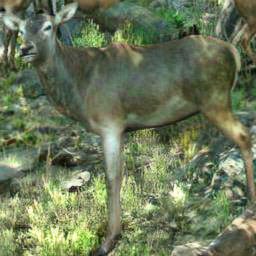} & 
\includegraphics[width=.172\linewidth,valign=m]{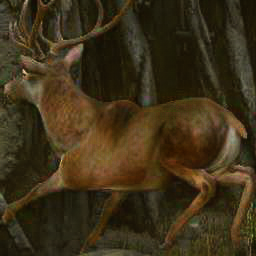} &
\includegraphics[width=.172\linewidth,valign=m]{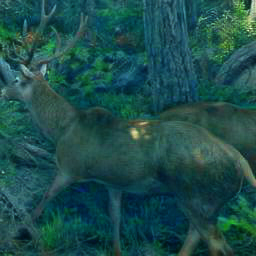} \\[+2ex]
\textit{real} & 
\includegraphics[width=.172\linewidth,valign=m]{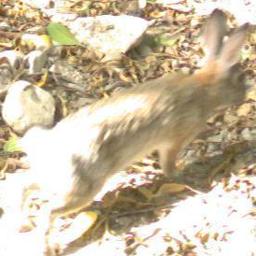} &
\includegraphics[width=.172\linewidth,valign=m]{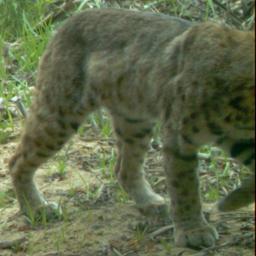} & 
\includegraphics[width=.172\linewidth,valign=m]{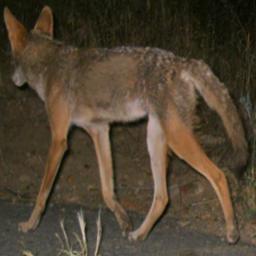} &
\includegraphics[width=.172\linewidth,valign=m]{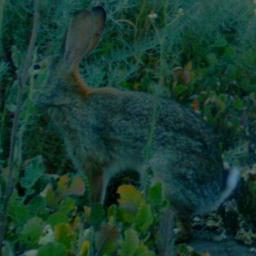} \\[+2ex]
\end{tabular}
\end{center}
\caption{\textbf{Examples generated by UNIT trained on the entire real training set of CCT-20.} When trained with all the categories as a target, the model learns to imitate the chromatic distribution of different locations seen during training. The outputs match with the style of images outside of the deer class.}
\label{fig:cct-20-model}
\end{figure*}

\begin{figure*}
\begin{center}
\centering
\begin{tabular}{ c c c c c }
\textit{syn} & \makecell{\textit{syn2real} \\ \textbf{CUT}}  & \makecell{\textit{syn2real} \\ \textbf{CycleGAN}}  & \makecell{\textit{syn2real} \\ \textbf{UNIT}}  & \textit{real}\\[+2ex]

\includegraphics[width=.12\linewidth,valign=m]{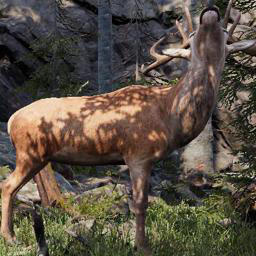} & \includegraphics[width=.12\linewidth,valign=m]{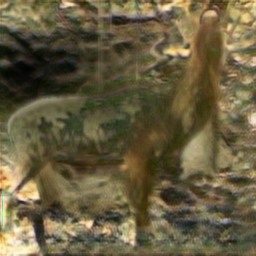} &
\includegraphics[width=.12\linewidth,valign=m]{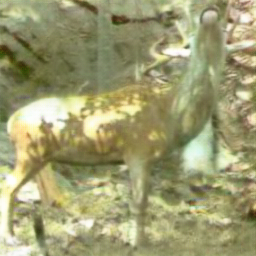} &
\includegraphics[width=.12\linewidth,valign=m]{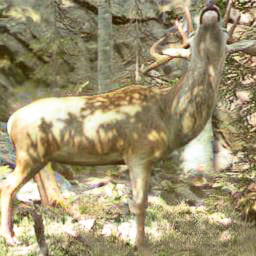} &
\includegraphics[width=.12\linewidth,valign=m]{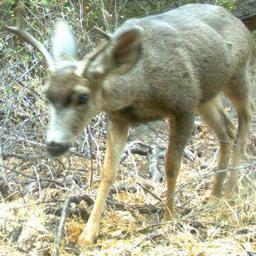}\\[+2ex]

\includegraphics[width=.12\linewidth,valign=m]{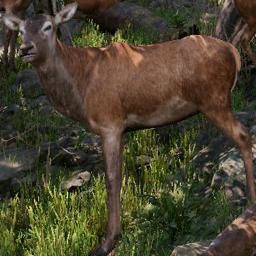} & \includegraphics[width=.12\linewidth,valign=m]{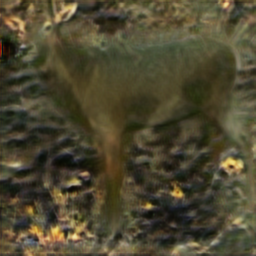} &
\includegraphics[width=.12\linewidth,valign=m]{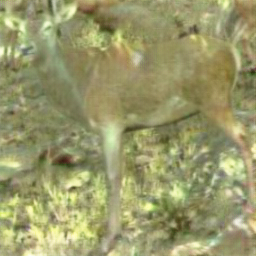} &
\includegraphics[width=.12\linewidth,valign=m]{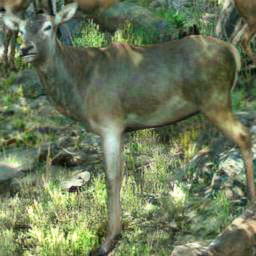} &
\includegraphics[width=.12\linewidth,valign=m]{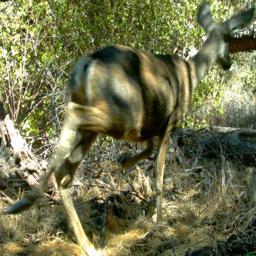}\\[+2ex]

\includegraphics[width=.12\linewidth,valign=m]{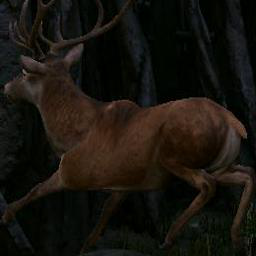} & \includegraphics[width=.12\linewidth,valign=m]{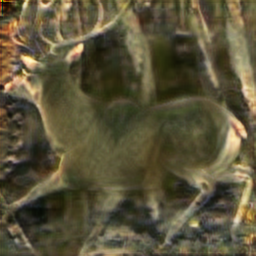} &
\includegraphics[width=.12\linewidth,valign=m]{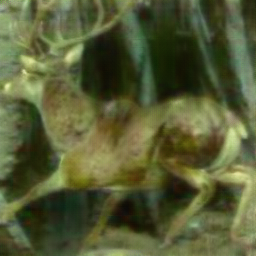} &
\includegraphics[width=.12\linewidth,valign=m]{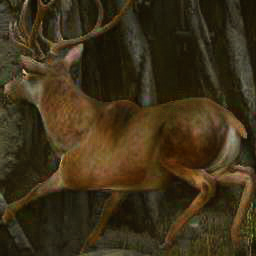} &
\includegraphics[width=.12\linewidth,valign=m]{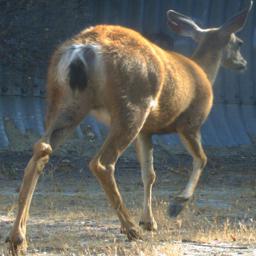}\\[+2ex]

\includegraphics[width=.12\linewidth,valign=m]{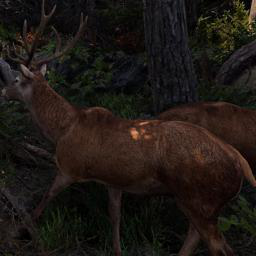} & \includegraphics[width=.12\linewidth,valign=m]{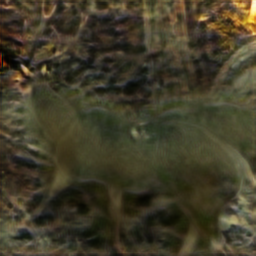} &
\includegraphics[width=.12\linewidth,valign=m]{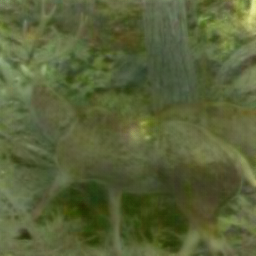} &
\includegraphics[width=.12\linewidth,valign=m]{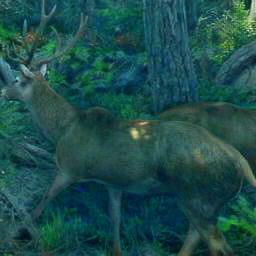} &
\includegraphics[width=.12\linewidth,valign=m]{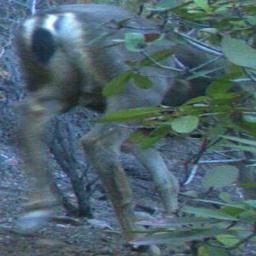}\\[+2ex]
\end{tabular}
\end{center}
\caption{\textbf{Comparison of the different unpaired I2I models trained on the same data.} These hand-picked examples show different outputs coming from the translation learned by the different models. UNIT shows better results, with more variance in the color distribution and sharpness in the refinement compared to CycleGAN and CUT.}
\label{fig:compare_models}
\end{figure*}

\begin{table}[h!]
\begin{center}
\begin{tabular}{lll}
\hline
\multicolumn{3}{c}{\textbf{CCT-deer}} \\
\hline
\textit{Correlation}                  & day  & night \\
\hline
correlation(syn, real)      & 0.73 & 0.36  \\
\textbf{correlation(syn2real, real)} & \textbf{0.96} & \textbf{0.96}  \\
correlation(syn2real, syn)  & 0.81 & 0.46  \\
\hline
\end{tabular}

\vspace*{0.25 cm}

\begin{tabular}{lll}
\hline
\multicolumn{3}{c}{\textbf{CCT-20}} \\
\hline
\textit{Correlation}               & day  & night \\
\hline
correlation(syn, real)      & 0.73 & 0.36  \\
\textbf{correlation(syn2real, real)} & \textbf{0.94} & \textbf{0.95}  \\
correlation(syn2real, syn)  & 0.70 & 0.29  \\
\hline
\end{tabular}
\end{center}
\caption{\textbf{Average color distribution correlations.} Measured between (i) the \textit{syn} and \textit{real} images, (ii) the \textit{syn2real} and \textit{real} images and (iii) the \textit{syn2real} and \textit{syn} images for both day and night. The model trained on CCT-20 (bottom) is performing similarly to the model trained on CCT-deer (top).}
\label{table:color_distribution}
\end{table}

\section{Experiments}
We use the entire collection of deer samples from CCT (denoted \textbf{CCT-deer}) to evaluate the different I2I translation models. The data is split between day (2342 samples) and night (3132 samples), and models are trained separately to translate bounding box crops resized to 256x256 pixels from the synthetic to the real domain.

To bridge the domain gap, three different unpaired I2I translation methods are compared. Using the official implementations of UNIT\footnote{https://github.com/mingyuliutw/UNIT}, CycleGAN\footnote{https://github.com/junyanz/pytorch-CycleGAN-and-pix2pix} and CUT\footnote{https://github.com/taesungp/contrastive-unpaired-translation}, we trained each model with the default hyperparameters on the same dataset. CycleGAN starts from a generative adversarial setting and adds a cycle consistency loss, to constraint the learned mapping \cite{CycleGAN2017}. Similarly, UNIT uses a shared-latent space constraint, separating the generator into an encoder and decoder component, enforcing cycle-consistency between those \cite{liu2017unsupervised}. CUT uses contrastive learning to encourage patches from the two domains to share mutual information \cite{CUT}. 

From a qualitative inspection of the learned translation of each model, UNIT produces qualitatively superior results (see Fig. \ref{fig:compare_models}). CycleGAN and CUT models appear to show less variance and sometimes introduce artifacts in their outputs. Because of this, we chose to use UNIT for the remainder of our experiments. 

The UNIT model trained on CCT-deer appears to visually imitate the locations seen during training from the real samples, altering the look of the synthetic image to mimic the real ones (see Fig. \ref{fig:deer_matrix}). Qualitatively, the model appears to learn to alter the colors of the image but the texture does not shift substantially. To measure this effect quantitatively, we analyze the color distribution and texture of the real data as well as the synthetic data pre- and post-translation.

\subsection{Color distribution}
\label{sub:color}

The most notable change in the translated samples is the shift in the color distribution of the synthetic samples, that appear to resemble the color scheme of the real samples. We consider day and night separately, as samples from the two are visually and statistically distinct.

\subsubsection{Day}
To evaluate the color difference for all samples obtained during the \textit{day}, we look at the sample-normalized distribution of the Hue value from the HSI colorspace, representing the pure color at each pixel regardless of saturation and illumination. To measure the distance between the \textit{real}, \textit{syn} and \textit{syn2real} distribution, we computed the Pearson correlation coefficient between each of them. The \textit{syn2real} correlation improves from 0.73 to 0.96 with \textit{real} samples and decreases from 1.0 to 0.81 with \textit{syn} samples (see Table \ref{table:color_distribution}).

\subsubsection{Night}
The \textit{night} samples are first converted to grayscale and their color features are captured by the sample-normalized distribution of pixel values. The \textit{syn2real} correlation improves from 0.36 to 0.96 with \textit{real} samples and decreases from 1.0 to 0.46 with \textit{syn} samples (see Table \ref{table:color_distribution}).\\

These measurements suggest that the model is able to approximate the distribution of the real samples for both \textit{day} and \textit{night}. An important observation is that the model imitates the color distribution of the \textbf{locations} that is trained on, uniformly altering the color of all the pixels in the image.

\begin{figure}[t!]
\begin{center}
\centering
\begin{tabular}{ c }
\includegraphics[width=1\linewidth,valign=m]{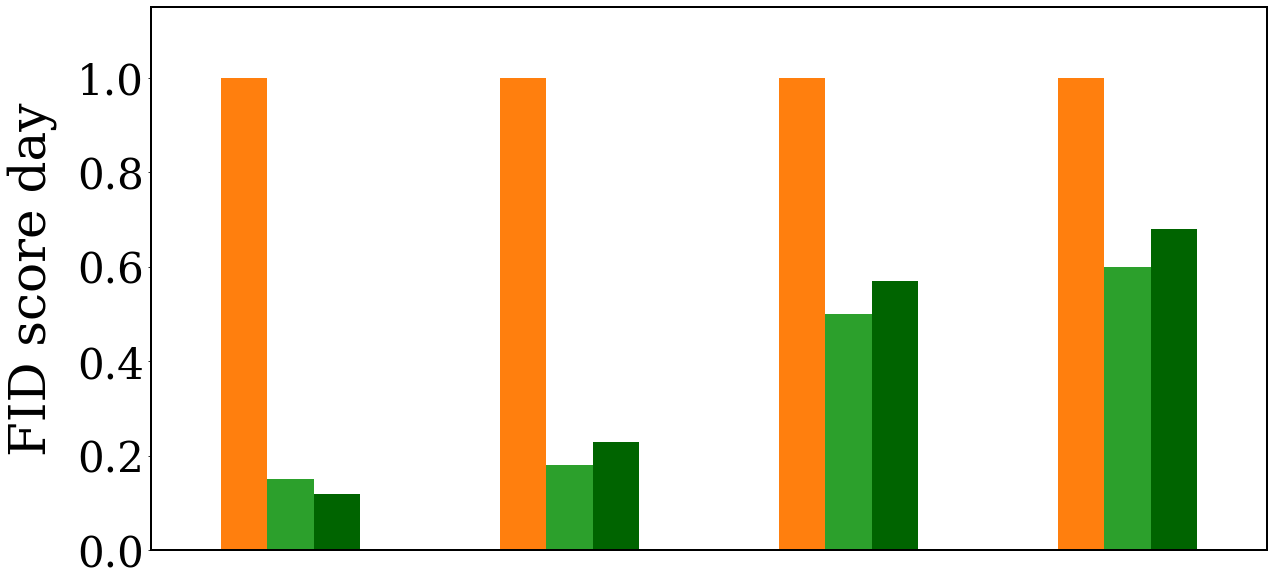}\\
\includegraphics[width=1\linewidth,valign=m]{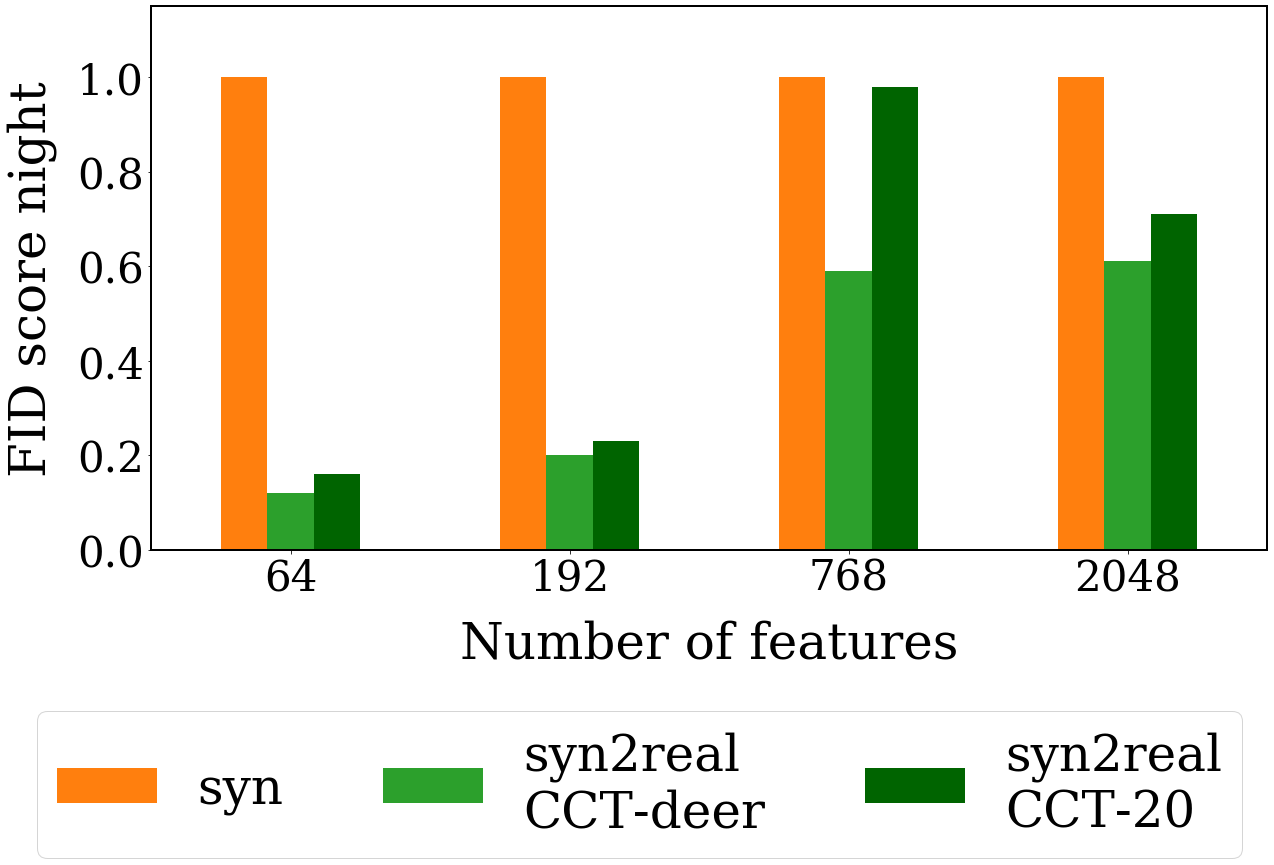} \\[+2ex]
\end{tabular}
\end{center}
\caption{\textbf{Normalized FID score computed for both \textit{day} and \textit{night} at different depths in feature space.} The score quantifies the distance from \textit{real} for both \textit{syn} and \textit{syn2real}. We express the \textit{syn2real} score as a fraction of the \textit{syn} at each architecture depth. We see that the gap is getting narrower at lower-level features.}
\label{fig:fid_scores}
\end{figure}

\subsection{Texture}

Another dimension through which we measure distance is texture space, quantifying the translation impact on the synthetic samples, compared to the real ones. To characterize textures, we use gray level co-occurrence matrix (GLCM) features \cite{glcm}. In particular, we extract \textit{contrast}, \textit{homogeneity}, \textit{energy} and \textit{entropy}.

The goal is to measure the difference between the texture of the fur of the animal across the two domains. To isolate this sub-experiment, a model is trained to map synthetic samples to a single location (location ID 34 in CCT) for which we have an abundant sample size during the day, allowing us to normalize the context in which textures are measured. This is due to the fact that texture changes across locations, due to lightning conditions and other factors. By translating to a single location, we can also manually pick real samples that were captured at a camera location with a similar look and confine the effect of the translation on the textures. Similar to \cite{peppers}, we manually crop 4 20x20 patches for 10 manually selected real and synthetic samples, compute and average GLCM features across the set.

The GLCM texture features measured on the \textit{syn2real} samples show a negligible improvement, compared to \textit{syn} samples. As confirmed by a qualitative inspection, the model is not considerably shifting the distribution of texture space. The positive delta introduced by the translation is small compared to the impact on the color distribution (see Section \ref{sub:color}).

\subsection{Exploring translation for a rare class}
\label{subsec:i2i_rare_class}
We have shown that it is possible to narrow the difference in visual appearance between \textit{syn} and \textit{real} camera trap data using CCT-deer as training set. Performing the same translation task for the deer class on the CCT-20 dataset becomes problematic due to the limited amount of target data (44 deer samples), but this represents a more realistic scenario for any rare species. That said, the previous experiment suggests that the mapping learned from the \textit{syn} to the \textit{real} data alters mostly the lower-level color features, with the textures being slightly changed. In other words, the model learns the appearance of the different \textbf{locations} presented during training. This suggests that the model could also learn a similar chromatic transformation using real images that do not necessarily correspond to the deer class, extending the target set from the 44 deer images to the 13,553 CCT-20 training images across all categories.

Using the entire CCT-20 training set as our target, the model replicates the chromatic distribution learned from the locations seen during training. As shown in Figure \ref{fig:cct-20-model}, those correspond to locations populated by categories outside of the \textit{deer} class. Using the same procedure described in Section \ref{sub:color} to measure color distributions, we find a correlation of 0.94 (day) and 0.95 (night) (see Table \ref{table:color_distribution}) with the real imagery.

\begin{figure}[t!]
\begin{center}
\includegraphics[width=1\linewidth]{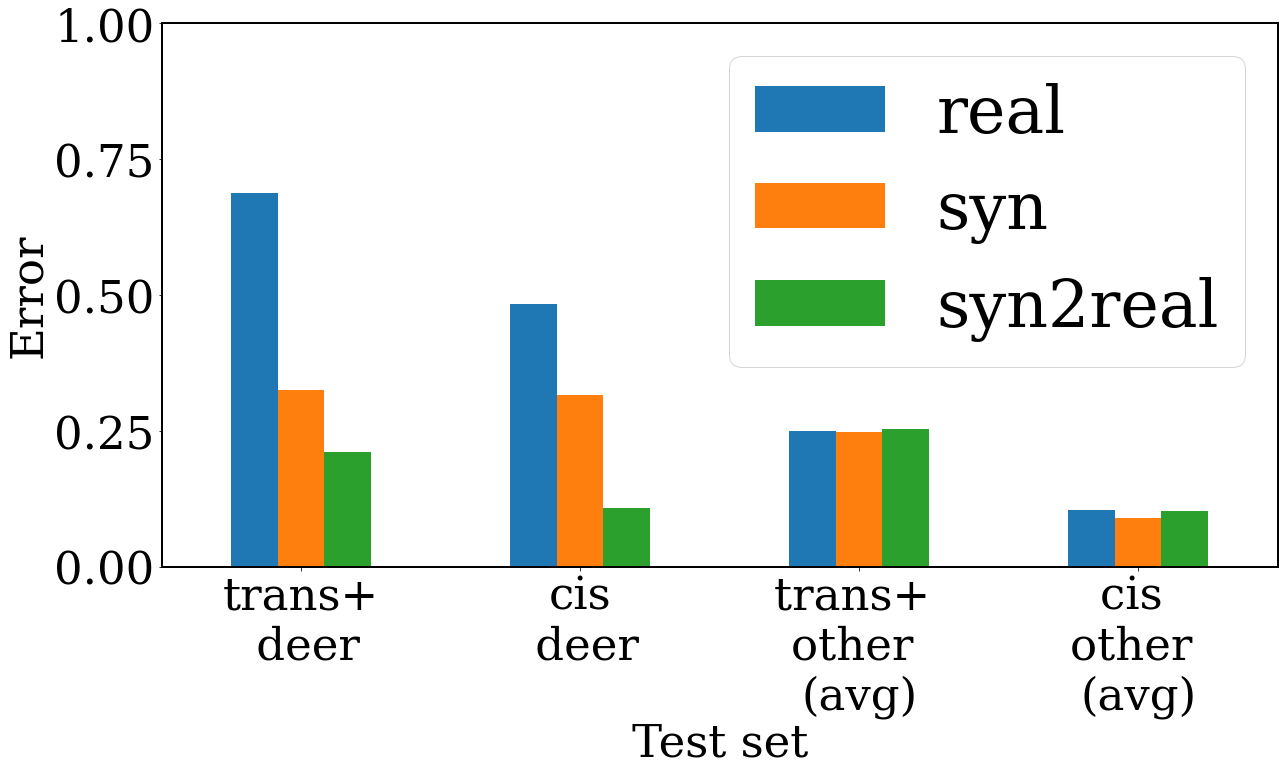}
\end{center} \caption{\textbf{Error rates measured on the classification of trans+ and cis test sets.} On both sets, the error rate for the "rare" deer class is significantly decreasing when the model in trained \textit{syn2real} data compared to just \textit{syn} samples. The change in the classification error of the other classes is negligible.}
\label{fig:classificaiton}
\end{figure}

\subsection{Feature space}
To further evaluate the quality of the two translation models, we use Fr\'echet Inception Distance (FID) to quantitatively capture domain similarity \cite{fid}. To capture the semantic distance at different architecture depths, activations of 64, 192, 768 and 2048 are extracted from a pretrained Inception classifier \cite{fid-code}.  

Figure \ref{fig:fid_scores} shows the computed FID between source and target for the respective feature dimensions. For both \textit{day} and \textit{night}, the translation method appears to close the gap most significantly early in the network, with the largest decrease at the first max-pooling layer (64 features), encapsulating lower-level features. The CCT-20 model performs similarly to the CCT-deer model, suggesting that the features corresponding to realness can be learned and transferred from a target set containing multiple categories, bypassing the need for large amounts of real data of our rare class.

\section{Classification}

The ultimate goal of our method is to improve classification of the rare class of interest by making our synthetic data more ``real". To test this, we finetune an Inception V3 model, pretrained on ImageNet, to classify species in bounding box crops from CCT-20. We use the same training parameters - learning rate, optimizer and input transformations - as  \cite{Beery_2020_WACV}. We compare classification results when training with (1) only the \textit{real} data, (2) augmenting it with 10K \textit{syn} samples (5K day, 5K night), and (3) augmenting with the same 10K synthetic samples post-translation  (\textit{syn2real}), using the model from \ref{subsec:i2i_rare_class}.\\

\noindent\textbf{Cis}\\
The cis test set is made up of held out images from camera locations seen during training. The error rate on cis test set decreases by 16\% from \textit{real} to \textit{syn} and improves by 37\% from \textit{real} to \textit{syn2real}. The model trained on \textit{syn2real} images improves the classification of the deer class on the cis test set by 21\%. The considerable improvement in the cis test set may stem from the ability of the I2I model trained on CCT-20 to mimic the low-level statistics of  each camera location.\\

\noindent\textbf{Trans}\\
The trans test set is composed of samples from camera locations not seen during training. This initial set is augmented with all the deer samples present in CCT (\textbf{trans+}). In the classification of deer for the trans+ test set, we see a 36\% decrease in error rate from \textit{real} to \textit{syn} and a 48\% decrease from \textit{real} to \textit{syn2real}. The 12\% improvement in the trans+ test set may stem from the training that is performed on translated samples that resemble the style of classes different from deer. This might help the model to generalize the classification to unseen locations for the rare deer class.\\

In both testing scenarios, we see a negligible change in the average error rate of the other classes ($\pm$$<$ 1\%) (see Fig. \ref{fig:classificaiton}).

\section{Conclusion}
The domain shift present in the low-level features between real and synthetic images can be effectively narrowed by simply imitating the color distribution of the locations in the target samples. Our experiments show this I2I translation can be learned using the entire training set of real samples, including samples from other categories. This is particularly beneficial when dealing with real-world long-tailed distributions, where rare classes are underrepresented. It remains to be tested how different I2I models deal with a multitude of domains (locations), investigating the distribution of the locations that the model is able to reproduce, compared to the training data.

The improvements on classification from the enhancement in ``realness" of the synthetic data is encouraging and could beneficially impact the wildlife monitoring of rare endangered species.

\section{Acknowledgements}
We would like to thank the USGS and NPS for providing data and Microsoft AI for Earth for providing compute resources. This work was supported in part by the Caltech Resnick Sustainability Institute and NSFGRFP Grant No. 1745301, the views are those of the authors and do not necessarily reflect the views of these organizations.

\newpage
{\small
\bibliographystyle{ieee_fullname}
\bibliography{main}
}

\end{document}